# Action Networks: A Framework for Reasoning about Actions and Change under Uncertainty

Adnan Darwiche and Moisés Goldszmidt
Rockwell Science Center
444 High Street, Suite 400
Palo Alto, CA 94301, U.S.A.
{darwiche, moises}@rpal.rockwell.com

## Abstract

This work proposes action networks as a semantically well founded framework for reasoning about actions and change under uncertainty. Action networks add two primitives to probabilistic causal networks: *controllable* variables and *persistent* variables. Controllable variables allow the representation of actions as directly setting the value of specific events in the domain, subject to preconditions. Persistent variables provide a canonical model of persistence according to which both the state of a variable and the causal mechanism dictating its value persist over time unless intervened upon by an action (or its consequences). Action networks also allow different methods for quantifying the uncertainty in causal relationships, which go beyond traditional probabilistic quantification. This paper describes both recent results and work in progress.

## 1 Introduction

The work reported in this paper is part of a project that proposes a decision support tool for plan simulation and analysis. The objective is to assist a human/computer planner in analyzing plan trade-offs and in assessing properties such as reliability, robustness, and ramifications under uncertain conditions. The core of this tool is a framework for reasoning about actions under uncertainty, called *action networks.* Action networks add two primitives to probabilistic causal networks (Bayes networks [12]): *controllable* variables and *persistent* variables. Controllable variables are the building blocks for the representation of actions in the domain. Persistent variables allow the modeling of time and change under uncertain conditions.

Controllable variables can be influenced directly by an agent. Thus, their value can be "set" regardless of the state and influences of actual possible causes in the domain. In this respect action networks follow the proposal in [8, 5] except for the introduction of the associated notion of a *precondition* for the action. Actions will be subject to preconditions connecting controllable variables to other variables that establish conditions for controllability.

At the heart of reasoning about actions lies the issue of modeling persistence and change: how and under what conditions should variables in a given domain persist over time when they are not influenced by actions? We propose a canonical model for persistence to dictate the states of special variables called persistent variables. Traditionally, the modeling of persistence has been accomplished by relating the state of a variable at time $t$ to its state at previous time-points. Problems with this approach has recently prompted researchers to explicitly model the causal mechanisms between variables in a network, and furthermore to persist the state of this mechanism over time (see Section 2.2). In this paper we further develop this model and propose a canonical model for the causal mechanisms. This model, called the suppressor model, is based on viewing a non-deterministic causal network as a parsimonious encoding of a more elaborate deterministic one in which suppressors (exceptions) of causal influences are explicated, and where all the uncertainty is in the state of the suppressors. The basic intuition here is that suppressors are believed to persist over time,[1] and that variables tend to persist when causal influences on them are deactivated by these suppressors.

Action networks employ quantified causal structures in the form of networks as a compact specification of a state of belief and as a formal language for specifying changes in a state of belief due to both observations and actions [12, 8, 5]. The causal structure allows us to deal with some of the key obstacles in reasoning about action and change such as, the frame and the concurrency problems, and reasoning about the indirect consequences of actions. To allow for different quantifications of the uncertainty in the causal rela-tions, an action network will consist of two parts: a directed graph representing a "blueprint" of the causal relationships in the domain and a quantification of these relationships. The quantification introduces a representation of the uncertainty in the domain because it specifies the degree to which causes will bring about their effects. Action networks will allow uncertainty to be specified at different levels of abstraction: point probabilities, which is the common practice in causal networks [12], order-of-magnitude probabilities, also known as $\varepsilon$-probabilities [8], and symbolic arguments, which allow one to explicate logically the conditions under which causes would bring about their effects [2]. In this paper, we will concentrate on order-of-magnitude probabilities as proposed in [8]. Other quantifications are described elsewhere [2, 5].

---
[1] The uncertainty of this persistence is determined by the specific domain.

This paper is organized as follows. Section 2 describes action networks. It starts with a brief review of network-based representations (Section 2.1), and continues with a description of the models of time and persistence (Section 2.2). Section 2.3 introduces the suppressor model. The representation of actions can be found in Section 2.4. Finally, Section 3 summarizes the main results and describes future work.

## 2  Action Networks

The specification of an action network requires three components: (1) the causal relations in the domain with a quantification of the uncertainty in these relations, (2) the set of variables that are "directly" controllable by actions with the variables that constitute their respective preconditions, and (3) variables that persist over time, which we will call *persistences* in this paper, and variables that do not persist over time, which we will call *events*.

Once the domain is modeled using this network-based representation (including uncertainty), an action network will be unfolded to create a more elaborate temporal network that includes additional nodes for representing actions and for representing the values of variables at different time points.

In this paper, variables will be denoted by lowercase letters $e$. Binary variables will be assumed to take values from $\{false, true\}$, which will be denoted by $e^-$ and $e^+$, respectively. For clarity of exposition, variables will be assumed to be binary, unless stated otherwise, and will be referred to as *propositions*. An instantiated proposition (or set of propositions) will be denoted by $\vec{e}$, and $\neg \vec{e}$ denotes the "negated" value of $\vec{e}$.

### 2.1  Network Representations

We briefly review some of the key concepts behind causal networks in this section given the central role they play in action networks. A *causal network* consists of a directed-acyclic graph $\Gamma$ and a quantification $Q$ over $\Gamma$. Nodes in $\Gamma$ correspond to domain variables, and the directed edges correspond to the causal relations among these variables. We denote the set of parents of a node $e$ in a belief network by $\pi(e)$. $\vec{\pi}(e)$ will denote a state of the propositions that constitute the parent set of $e$. The set conformed by $e$ and its parents $\vec{\pi}(e)$ is usually referred to as the causal family of $e$. Figure 1 depicts the causal family of *engine_running*. This network in conjunction with its quantification in terms of the $\kappa$-calculus depicted in Table 1 represents the belief that the engine will be running given that we turn the ignition key.[2]

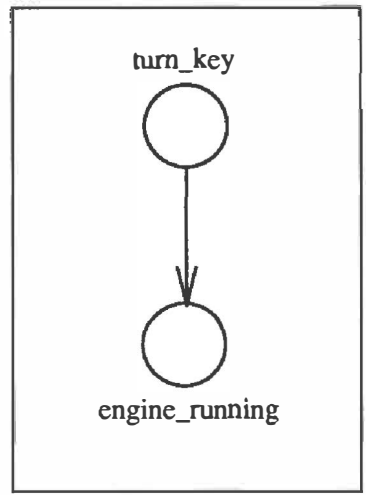

Figure 1: Causal family for *engine_running*, representing that *turn_key* causally influences *engine_running*.

| turn_key | engine_running$^+$ | engine_running$^-$ |
|---|---|---|
| true | 0 | 2 |
| false | 0 | 0 |

Table 1: $\kappa$-Quantification for the *engine_running* family. It represents the default causal rule "If *turn_key* then *engine_running*".

The quantification of $\Gamma$ over the families in the network encodes the uncertainty in the causal influences between $\pi(e)$ and $e$. In Bayesian networks, this uncertainty is encoded using numerical probabilities [12]. There are, however, other ways to encode this uncertainty that do not require an exact and complete probability distribution. Two recent approaches are the $\kappa$-calculus where uncertainty is represented in terms of plain beliefs and degrees of surprise [8], and argument calculus where uncertainty is represented using logical sentences as arguments [2]. These approaches are regarded as abstractions of probability theory since they retain the main properties of probability including Bayesian conditioning [4, 9].

---

[2] Appendix A reviews the main ideas behind the $\kappa$-calculus.

An important property of these networks is that a complete and coherent state of belief can be reconstructed from the local quantifications of the families. Thus, they constitute a compact specification of a state of belief. In probabilities for example, given a network containing nodes $x_1, \ldots, x_n$,

$$P(\vec{x}_1, \ldots, \vec{x}_n) = \prod_{1 \leq i \leq n} P(\vec{x}_i | \vec{\pi}(x_i)) \qquad (1)$$

Similar equations can be obtained for the $\kappa$-calculus and for argument calculus. Since, in this paper we concentrate on a quantification based on plain beliefs using kappa rankings, we provide a brief review of their main properties in Appendix A.

## 2.2 Time and Persistence

When unfolding a persistent variable in an action network, new variables are added to represent its values at different time points. This leads to a more elaborate causal network that spans over time. The structure of this temporal network is the focus of this section.

Action networks appeal to two assumptions, the realization of which lead to a specific proposal of how to expand an action network into a temporal causal network. The first assumption states that the causal relations between variables at a specific time point are similar to those explicated in the action network. That is, if $e$ has causes $c_1, \ldots, c_n$ in the action network, it will have these causes at every time point. The second assumption in action networks relates to temporal persistence. It says that the state of the system modeled by an action network persists over time (with a certain degree of uncertainty) in the absence of external intervention. In the remainder of this section we formalize a proposal that realizes these assumptions.

Before we present our persistence model though, it will be illustrative to discuss two intermediate proposals that have inspired the current one.

### Persisting Variable States

Our first approach required that we make each persistence at time $t$ a direct cause of itself at time $t+1$. This was intended to represent the influence that the past state of a persistence has on its immediate future state. For example, assuming that *turn_key* is an event and *engine_running* is a persistence, this proposal leads to Figure 2. This approach is reminiscent of a number of proposals in the literature [13, 6]. It fails, however, to capture the notion of persistence that we are after because it leads to conclusions that are weaker than one would expect. For example, assume that the probability of *engine_running*$^+$ at time $t$ given *turn_key*$^+$ at time $t$ and *engine_running*$^-$ at $t-1$ is .9. Suppose now that we turn the key at time 0 but the engine does not start. We repeat the experiment at times $1, \ldots, n-1$ with similar results (i.e., the engine does not run). In this model of persistence, the probability of *engine_running*$^+$ at time $n$ is still .9 given *turn_key*$^+$ at time $n$. Yet, intuitively, we would expect the car not to start at time $n$ given the previous sequence of observations.[3]

### Persisting Causal Mechanisms

The previous example suggests that it is not enough to persist the state of *engine_running*. One must also persist the causal mechanism between *turn_key* and *engine_running* [11]. The reason why we expect the car not to start is due to our previous observations which lead us to conclude that the causal mechanism between *turn_key* and *engine_running* is not behaving normally. Moreover, we seem to assume that the state of the existing mechanism persists over time since no one intervened to change it. One way to capture these intuitions is to explicitly provide a representation of the causal mechanism between an event $e$ and its causes $\pi(e)$ in the network. This solution requires that we add (at least) another parent $\mathcal{U}(e)$ to each family, which represents all possible causal mechanisms between $\pi(e)$ and $e$. The node $e$ will then be *deterministic* since its state will functionally depend on the state of $\pi(e)$ and $\mathcal{U}(e)$. This model is intuitively appealing in that it encodes the causal relation of a family as a set of functions between the direct causes $\pi(e)$ and their effect $e$, where the state $\mathcal{U}(e)$ selects the "active" function that specifies the current causal relation. The likelihood that any of these functions is active depends on the likelihood of the state of the variable $\mathcal{U}(e)$.[4]

Using this approach we persist the functional mechanism represented by the $\mathcal{U}(e)$ nodes in each family, as shown in Figure 3 with regards to the *engine_running* family.[5]

---

[3] This problem will re-appear even if more refined models of the domain are proposed. One could, for example, add more causal parents representing the exceptions that would prevent the engine from running given that the key is turned. One such exception can be a *dead_battery*. Although a step in the right direction, such refinements will not solve the problem above, since we can always reproduce the counter-example by introducing the appropriate set of observations (e.g., the battery was OK at each point in time, including time $n$).

[4] The assumption behind this representation is that the uncertainty recorded in the quantification of each family in a network $\Gamma$ expresses the incompleteness of our knowledge in the causal relation between $e$ and its set of direct causes $\pi(e)$. This incompleteness arises because $e$ interacts with its environment in a complex manner, and this interaction usually involves factors which are exogenous to $\pi(e)$. Furthermore, these factors are usually unknown, unobservable or too many to enumerate. Thus, we can view a non-deterministic causal family as a parsimonious representation of a more elaborate, deterministic causal family, where the quantification summarizes the influence of other factors on $e$.

[5] Similar representations were used by Pearl and Verma [14] for discovering causal relationships from observations, and by Druzdzel and Simon [7] in their study about the representation of causality in Bayes networks.

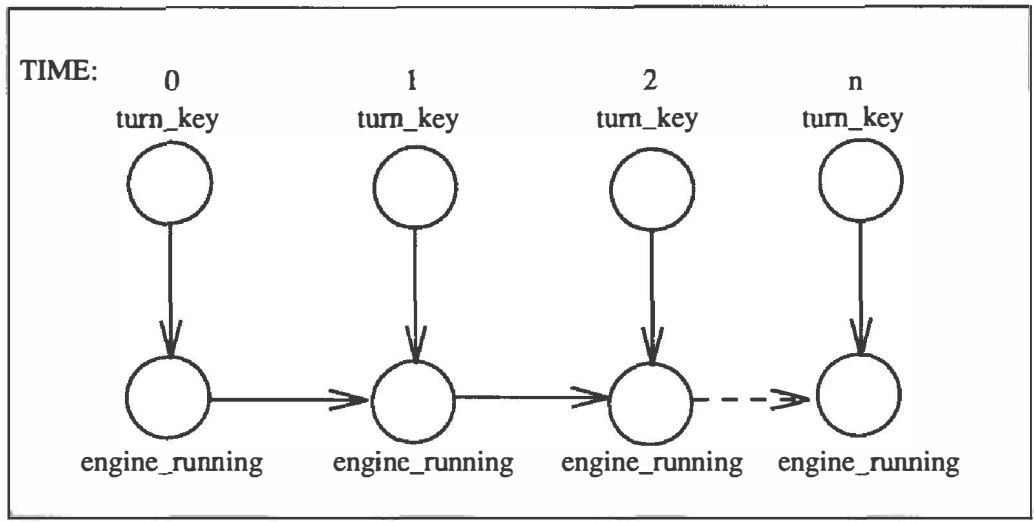

Figure 2: "Simple" temporal expansion of the family encoding the relation between *turn_key* and *engine_running*. Proposition *engine_running* is taken to be a *persistence* while proposition *turn_key* is assumed to be an *event*.

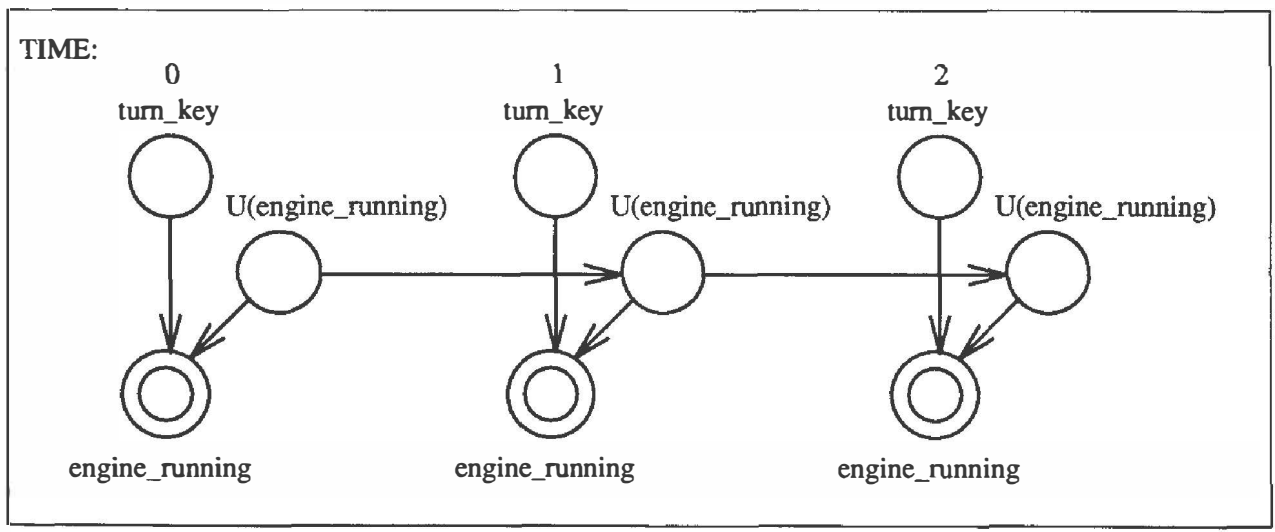

Figure 3: Temporal expansion of the family encoding the relation between *turn_key* and *engine_running*, and including the dynamics of their causal relation through the persistence of the node $\mathcal{U}(engine\_running)$. The concentric circles indicate that the *engine_running* node is deterministic, and depends functionally on the states of *turn_key* and $\mathcal{U}(engine\_running)$.

Unfortunately, even though the model in Figure 3 explicates and persists the causal mechanism between causes and their effects, it is too weak to capture the notion of persistence we are after. Suppose for example that we turn the key at time 0. The system will then infer that the engine will be running at time 0 with probability .9. However, the model will not be able to conclude that the engine will continue to be running at times 1, 2, ..., and so on. In fact, from the topology of Figure 3, we can see that whether the engine is running at time $t+1$ is marginally independent of whether the key was turned at time $t$, which is contrary to what we would expect from the persistence assumption.

### Persisting Variable States and Causal Mechanisms

The approach we adopt, depicted in Figure 5, can be regarded as a combination of the temporal networks in Figures 2 and 3. The proposition *engine_running* is functionally determined by *turn_key*, $S(engine\_running)$, and $I(engine\_running)$. The variable $S(engine\_running)$ captures all possible suppressors of the causal influences that the proposition *turn_key* has on *engine_running*. The variable $I(engine\_running)$ decides the state of *engine_running* when the the suppressors manage to deactivate the causal influence of *turn_key* on *engine_running*, and it is directly influenced by the past state of *engine_running*.

In the static case, when time is not involved, the proposal can be viewed as splitting the variable $\mathcal{U}(e)$ into two variables, $S(e)$ and $I(e)$. Note, however, that once we expand over time the notion of a causal mechanism has a broader scope because it has to account for the previous value of proposition $e$. The semantics of the variable $S(e)$ assumes that the uncertainty in the causal relation between $e$ and its causes $\pi(e)$ is due to a set of abnormalities and exceptions that suppress this causal influence. When this influence is suppressed due to these exceptions and abnormalities the value of $e$ is set according to its previous state (represented by the variable $I(e)$). This model of persistence makes two assumptions. First, it assumes that the state of suppressors tend to persist over time (with a degree of uncertainty determined by the specific application). Second, it assumes that the state of variable $I(e)$ is determined by the state of $e$ at the previous time point.[6]

This model is not only intuitive and solves the problems outlined above, but it allows for a modular quantification of the network: the uncertainty in the causal relations, the uncertainty in the persistence of suppressors, and the uncertainty in the persistence of variables

---

An expansion similar to the one in Figure 3 is used by Balke and Pearl [1] for answering probabilistic counterfactual queries, and by Heckerman and Shachter [11] for capturing the notion of causal persistence.

[6] Both these assumptions can be relaxed and lead to more elaborate models (see Section 3).

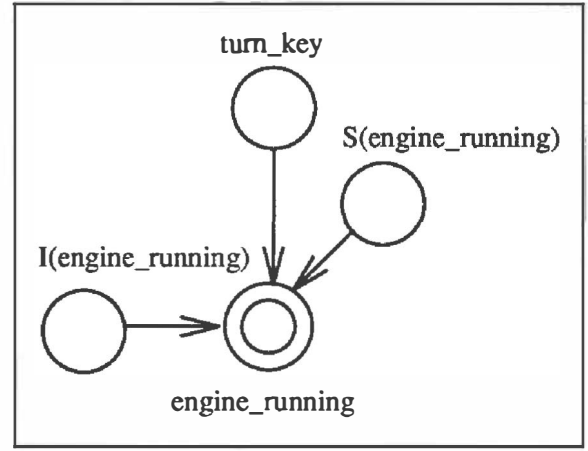

Figure 4: The suppressor model as a functional expansion of causal network.

can be specified independently (see Eqs. 5 and 6).

The following section will discuss the suppressor model in more detail.

### 2.3 The Suppressor Model

To formally describe the suppressor model we first examine how it expands a "static" causal network into a functional one, where all causal relations are deterministic and all the uncertainty is about the states of root nodes. Then we show how this functional expansion of a causal network lends itself naturally for capturing the persistence assumptions that we stated in the previous section.

As a proposal for functionally expanding a causal network, the suppressor model is based on the following intuition. The uncertainty in the causal influences between $\pi(e)$ and $e$ is a summary of the set of exceptions that attempt to defeat or suppress this relation. For example, "a banana in the tailpipe" is a possible suppressor of the causal influence that *turn_key* has on *engine_running*. The expansion into the suppressor model makes the uncertain causal relation between $\pi(e)$ and $e$ functional by adding a new parent $S(e)$ to the family, which corresponds to the suppressors to the causal relation. In addition to $S(e)$, another parent $I(e)$ is added, which will set the state of $e$ in those cases in which the suppressors manage to defeat the causal influence of $\pi(e)$ on $e$. In these cases we say that the suppressors are "active". Figure 4 depicts the expansion of the *engine_running* family.

Once a causal network is functionally expanded according to the suppressor model, the persistence assumption stated in the previous section can be formalized by taking the variable $I(e)$ to represent the previous state of $e$ — see Figure 5. The intuition being that in those cases where the suppressors manage to prevent the natural causal influences on $e$, the state

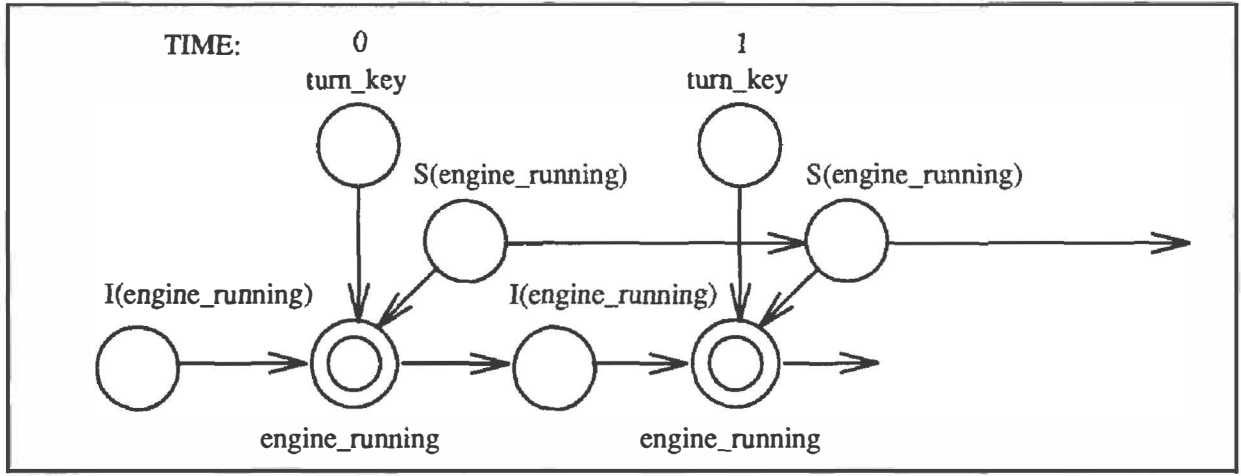

Figure 5: The suppressor model for the *engine_running* family.

of $e$ should simply persist and follow its previous state.

We will now present the suppressor model formally. Let $S(e)$ take values out of the set $\{\omega^0, \omega^1, \omega^2, \ldots\}$[7] where $S(e) = \omega^s$ stands for "a suppressor of strength $s$ is active." The function **F** relating $e$ to its direct causes $\pi(e)$, the suppressors $S(e)$, and the variable $I(e)$ is given by

$$\mathbf{F}(\vec{\pi}(e), \omega^i, \vec{I}(e)) = \begin{cases} e^+, & \text{if } \kappa(e^- \mid \vec{\pi}(e)) > i; \\ e^-, & \text{if } \kappa(e^+ \mid \vec{\pi}(e)) > i; \\ \vec{I}(e), & \text{otherwise.} \end{cases} \quad (2)$$

Where $\kappa(e|\pi(e))$ represents the strength of belief in the causal relation between $\pi(e)$ and $e$. Eq. 2 says that if the strength of the active suppressor $\omega^i$ is less than the causal influence of $\pi(e)$ on $e$, then the state of $e$ is dictated by the causal influence. Otherwise, the suppressor is successful, the causal influence is suppressed, and the state of $e$ is the same as the state of $I(e)$.

The translation of **F** into a $\kappa$ matrix is given by below:

$$\kappa'(\vec{e} \mid \vec{\pi}(e), \omega^i, \vec{I}(e)) = \begin{cases} 0, & \text{if } \vec{e} = \mathbf{F}(\vec{\pi}(e), \omega^i, \vec{I}(e)); \\ \infty, & \text{otherwise.} \end{cases} \quad (3)$$

The prior distribution of beliefs on $S(e)$ is given by

$$\kappa'(\omega^i) = i, \quad (4)$$

which reflects the intuition that suppressors are typically inactive, and that the stronger the suppressor is, the more unlikely that it will be active.

Using the suppressor model, we can take any non-deterministic network quantified with kappas and automatically expand it into a functional network in the

---

[7]In general, the suppressor takes values in $\{\omega^0, \omega^1, \omega^2, \ldots, \omega^\infty\}$. In practice, however, it suffices for the suppressor to take values in $\{\omega^k\}$ where the ranking $k$ appears in the matrix of $e$.

sense that all causal relations are deterministic and the only uncertainty is regarding suppressors. Moreover, we get the following guarantee about the resulting functional network, which says that the new network captures all the information which the initial network was set to capture. Let $\kappa$ represent the quantification of a non-deterministic network and let $\kappa'$ represent the quantification of its functional expansion:

**Theorem 1** $\kappa(\vec{e} \mid \vec{\pi}(e)) = \kappa'(\vec{e} \mid \vec{\pi}(e))$.

The proof of this theorem relies on marginalizing $\kappa'(\vec{e}|\vec{\pi}(e), \vec{S}(e), \vec{I}(e))$ over all the states of $S(e)$ and $I(e)$.

Table 2 shows the automatic functional expansion of the causal relation between *engine_running* and $\pi(engine\_running)$ (depicted in Figure 1 and $\kappa$-quantified in Table 1) reflecting Eq. 2.

| turn_key | S(engine_running) | I(engine_running) | engine_running |
|---|---|---|---|
| true | $\omega^0$ | true | true |
| true | $\omega^0$ | false | true |
| true | $\omega^2$ | true | true |
| true | $\omega^2$ | false | false |
| false | $\omega^0$ | true | true |
| false | $\omega^0$ | false | false |
| false | $\omega^2$ | true | true |
| false | $\omega^2$ | false | false |

Table 2: Deterministic function relating *engine_running* and its parents in the suppressor model.

In order to complete the temporal expansion, shown in Figure 5, we must quantify the uncertainty on the persistence arcs. Causal families are connected, across time points, through the $I(e)$ node and the suppressor $S(e)$ node. The conditional beliefs $\kappa(S(e_{t+1})|S(e_t))$ and $\kappa(\vec{I}(e_{t+1})|\vec{e_t})$ will formally determine the strength of persistence across time. Both conditional beliefs will encode a bias against a change of state, which captures the intuition that any change of state must be causally motivated. Note that this quantification is done modularly and independently of the quantification of the uncertainty in the causal relations. This separation is important for fast and efficient model building.

The quantification of these beliefs will be of course tied directly with an actual application and a specific domain. In our experiments, and for the planning domain implemented we had intuitive results with the following model in which the strength of the persistence assumption is proportional to the strength of the change in the state of the suppressor:

$$\kappa(\omega_{t+1}^i|\omega_t^j) = |j - i|. \tag{5}$$

The assumption of persistence for the $I(e)$ node corresponds to the following equation:

$$\kappa(\vec{I}(e_{t+1})|\vec{e_t}) = \begin{cases} p, & \text{if } \vec{I}(e_{t+1}) \neq \vec{e_t}; \\ 0, & \text{otherwise.} \end{cases} \tag{6}$$

Since $I(e_t)$ determines the state of $e_{t+1}$ when suppressors are active, the number $p$ can be interpreted as the degree of surprise in a non-causal change of state of the proposition $e_{t+1}$.[8]

**Example.** Consider the *engine_running* network in Figure 1 quantified as in Table 1. Assume that this network is temporally expanded using the suppressor model (see Eq. 3 and Figure 5). Given the ignition key is turned at times $1, 2, 3, \ldots, n-1$ and that the engine is not running at times $1, 2, 3, \ldots, n-1$, the model will yield the belief that the engine will not be running at time $n$, given that the key is turned at time $n$. On the other hand, given that the key is turned at time 0, the system will infer that the engine will be running at time 0, and moreover that it will be running at times $1, \ldots, n$.

## 2.4 Actions and Preconditions

For the representation of actions, we essentially follow the proposal in [8], which treats actions as external direct interventions that deterministically set the value of a given proposition in the domain. Actions are specified by indicating which nodes in the causal network are controllable and under what preconditions. Syntactically, we introduce a new node " ■ " denoting controllability. In Figure 6 for example, both *fired_gun* and *loaded_gun* are controllable propositions. A suitable precondition for both nodes can be *holding_gun*, which can be represented as just another direct cause of these nodes. The corresponding matrices will then be constructed to reflect the intuition that the action $do_e$ will be effective only if the precondition is true; otherwise, the state of a node $e$ is decided completely by the state of its natural causes (that is, excluding $do_e$ and the preconditions of $do_e$). Let the variable $do_e$ take the same values as $e$ in addition to the value

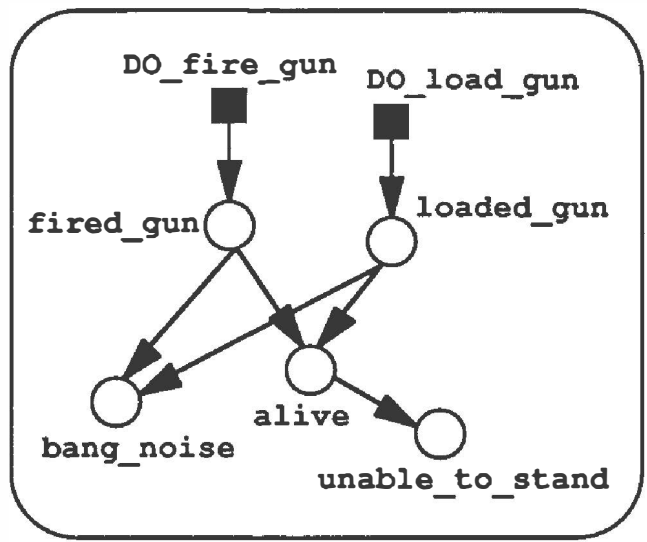

Figure 6: Causal family containing actions and ramifications (YSP).

*idle*.[9] The new parent set of $e$ after $e$ is declared as controllable will be $\pi(e) \cup \{do_e\}$. The new ranking $\kappa'(\vec{e}|\vec{\pi}(e) \wedge \vec{do_e})$ is

$$\kappa'(\vec{e}|\vec{\pi}(e) \wedge \vec{do_e}) = \begin{cases} \kappa(\vec{e}|\vec{\pi}(e)) & \text{if } \vec{do_e} = idle \\ \infty & \text{if } \vec{do_e} \neq \vec{e} \\ 0 & \text{if } \vec{do_e} = \vec{e} \end{cases} \tag{7}$$

For simplicity of exposition we have omitted possible preconditions. Their inclusion will just involve a refinement of the cases in Eq. 7 to reflect the fact that an action is possible iff its preconditions are satisfied.

The advantage of using this proposal as opposed to others, such as STRIPS, is that the approach based on direct intervention takes advantage of the network representation for dealing with the indirect consequences of actions and the related frame problem. In specifying the action "shooting", for example, the user need not worry about how this action will affect the state of other related consequences such as *bang_noise* or *alive* (see Figure 6).

**Example.** Consider the example in Figure 6 encoding a version of the Yale Shooting Problem (YSP) [10]. The relevant piece of causal knowledge available is that if a victim is shot with a loaded gun, then she/he will die. There are two possible actions, shooting and loading/unloading the gun. It is also assumed that both loaded and alive persist over time. Given this information the implementation of action networks will expand the network in Figure 6 both functionally and temporally.

In the first scenario, we observe at time 0 that the individual is alive and that the gun is loaded, and that there is a shooting action at time 2. The model will yield that alive at time 2 will be false.[10] This scenario shows the interplay between the persistences and the causal influences in the network.

---

[8] This value does not need to be constant, although it will assumed to be so in the remainder of the paper.

[9] Thus, if $e$ is binary $\vec{do_e} \in \{e^+, e^-, idle\}$.

In the second scenario, it is observed that at time 2 alive is true (the victim actually survived the shooting). The model will then conclude that: first, the gun must have been unloaded prior to the shooting (although it is not capable of asserting when), and furthermore, the belief of an action leading to unloading the gun increases (proportional to the degree of persistence in loading). This scenario displays the model capabilities for performing abductive reasoning including reasoning about the set of actions that would yield a given observation.

## 3  Conclusions and Future Work

We described the models of time, persistence, and action that constitute the core of action networks as a formalism for reasoning about actions and change under uncertainty. The notion of persistence was formalized through the suppressor model which evolved from other proposals for extending causal networks over time. The suppressor model should be viewed as one canonical model for representing persistence. Relaxing the assumptions in this model will yield other possible, more complex representations. For example we can make the $I(e)$ node depend on more than one past instance of $e$. This would allow the representation of time-decaying functions for the dependence of $e$ on its past values. We are currently exploring and characterizing this and other alternatives with the objective to provide a library of such models that would assist the user in encoding the dynamics of causal relations in the domain of interest.

We also intend to add notions of utility and preferences on outcomes and to explore the use of action networks in the formulation of a plan, given a set of objectives. The paths we are currently exploring include abductive methods for uncovering the sequence of actions that can lead to a specific set if beliefs, and the possibility of interfacing action networks as an evaluation component to a planning module.

This paper has focused on the $\kappa$-calculus instantiation of action networks. Future work includes allowing other quantifications of uncertainty, such as probabilities, and arguments, and even a mixture of these. We are also studying a probabilistic (and argument-based) interpretation of the suppressor model. The first steps toward the quantification of action networks with arguments is reported in [5].

Finally we remark that all the features of action networks described in this paper, including the suppressor model expansion, the temporal expansion, and the specification of actions, are fully implemented on top of CNETS [3].[11] All the examples described in this paper were tested using this implementation.

## Acknowledgments

We wish to thank P. Dagum for discussions on the nature and properties of the functional expansion of a causal network. We also thank C. Boutilier, J. Pearl, Y. Shoham and the Nobotics group at Stanford for discussions on the representation of actions. D. Draper, D. Etherington, and D. Heckerman provided useful comments on a previous version of this paper.

This work was partially supported by ARPA contract F30602-91-C-0031, and by IR&D funds from Rockwell Science Center.

## A  Appendix: A Review of The Kappa Calculus.

We provide a brief summary of the $\kappa$-calculus and how it can quantify over the causal relations in a network $\Gamma$.

Let $\mathcal{M}$ be a set of worlds, each world $m \in \mathcal{M}$ being a truth-value assignment to a finite set of atomic propositional variables $(e_1, e_2, \ldots, e_n)$. Thus any world $m$ can be represented by the conjunction of $\vec{e_1} \wedge \ldots \wedge \vec{e_n}$. A belief *ranking function* $\kappa(m)$ is an assignment of non-negative integers to the elements of $\mathcal{M}$ such that $\kappa(m) = 0$ for at least one $m \in \mathcal{M}$. Intuitively, $\kappa(m)$ represents the degree of surprise associated with finding a world $m$ realized, worlds assigned $\kappa = 0$ are considered serious possibilities, and worlds assigned $\kappa = \infty$ are considered absolute impossibilities. A proposition $\vec{e}$ is *believed* iff $\kappa(\neg \vec{e}) > 0$ (with degree $k$ iff $\kappa(\neg \vec{e}) = k$), where

$$\kappa(\vec{e}) = \min_{m \models \vec{e}} \kappa(m). \tag{8}$$

$\kappa(m)$ can be considered an order-of-magnitude approximation of a probability function $P(m)$ by writing $P(m)$ as a polynomial of some small quantity $\varepsilon$ and taking the most significant term of that polynomial, i.e., $P(m) \cong C\varepsilon^{\kappa(m)}$. Treating $\varepsilon$ as an infinitesimal quantity induces a conditional ranking function $\kappa(e|c)$ on propositions and wffs governed by properties derived from the $\varepsilon$-probabilistic interpretation [9].

A causal structure $\Gamma$ can be quantified using a ranking belief function $\kappa$ by specifying the conditional belief for each proposition $\vec{e}$ given every state of $\vec{\pi}(e)$, that is $\kappa(\vec{e}|\vec{\pi}(e))$. Thus, for example, Table 1 shows the $\kappa$-quantification of the *engine_running* family in Figure 1. Table 1 is called the "$\kappa$-matrix" for the *engine_running* family. It represents the default causal rule: "If $turn\_key^+$ then $engine\_running^+$".

---

[10] The reasons for this conclusion are due to the conditional independences assumed in the causal network representation. They are formally explained in depth in [12][Chapter 10] and [8].

[11] CNETS is an experimental environment for representing and reasoning with generalized causal networks, which allow the quantification of uncertainty using probabilities, $\kappa$ degrees of belief, and logical arguments [2].

The $\kappa$-calculus does not require commitment about the belief in $e^+$ or $e^-$. Thus, as Table 1 indicates, we may be ignorant about the status of the *engine_running* given that the ignition key is not turned. In such a case the user may specify that both $\kappa(engine\_running^+|turn\_key^-) = \kappa(engine\_running^-|turn\_key^-) = 0$ indicating that both alternatives are plausible.

Once similar matrices for each one of the families in a given network $\Gamma$ are specified, the complete ranking function can be reconstructed by requiring that

$$\kappa(m) = \sum_i \kappa(\vec{e_i}|\vec{\pi}(e_i)), \qquad (9)$$

and queries about belief for any proposition and wff can be computed using any of the well-known distributed algorithms for belief update [12, 3]. The class of ranking functions that comply with the requirements of Eq. 9 for a given network $\Gamma$ are called *stratified rankings*. These rankings present analogous properties about conditional independence to those possessed by probability distributions quantifying a network with the same structure [8].